\documentclass[runningheads]{llncs}
\usepackage[T1]{fontenc}
\usepackage{graphicx}
\usepackage{hyperref}
\begin{document}
\title{ChatGPT-4 as a Tool for Reviewing Academic Books in Spanish}

\author{Jonnathan Berrezueta-Guzman\inst{1}\orcidID{0000-0001-5559-2056} \and
Laura Malache-Silva\inst{2}\orcidID{0009-0004-2846-8710} \and Stephan Krusche\inst{1}\orcidID{0000-0002-4552-644X}}

\authorrunning{Berrezueta-Guzman et al.}

\institute{Technical University of Munich, Germany 
 \and
CEDIA, Ecuador\\
\email{s.berrezueta@tum.de}, 
\email{laura.malache@cedia.org.ec},
\email{krusche@tum.de}}

\maketitle              

\begin{abstract}
This study evaluates the potential of ChatGPT-4, an artificial intelligence language model developed by OpenAI, as an editing tool for Spanish literary and academic books. The need for efficient and accessible reviewing and editing processes in the publishing industry has driven the search for automated solutions. ChatGPT-4, being one of the most advanced language models, offers notable capabilities in text comprehension and generation. In this study, the features and capabilities of ChatGPT-4 are analyzed in terms of grammatical correction, stylistic coherence, and linguistic enrichment of texts in Spanish. Tests were conducted with 100 literary and academic texts, where the edits made by ChatGPT-4 were compared to those made by expert human reviewers and editors. The results show that while ChatGPT-4 is capable of making grammatical and orthographic corrections with high accuracy and in a very short time, it still faces challenges in areas such as context sensitivity, bibliometric analysis, deep contextual understanding, and interaction with visual content like graphs and tables. However, it is observed that collaboration between ChatGPT-4 and human reviewers and editors can be a promising strategy for improving efficiency without compromising quality. Furthermore, the authors consider that ChatGPT-4 represents a valuable tool in the editing process, but its use should be complementary to the work of human editors to ensure high-caliber editing in Spanish literary and academic books.
\let\thefootnote\relax\footnote{Preprint. Paper accepted in the 18\textsuperscript{th} Latin American Conference on Learning Technologies (LACLO 2023).}

\end{abstract}

\section{Introduction}

ChatGPT-4, developed by OpenAI, is an advanced iteration of artificial intelligence language models. 
It utilizes extensive datasets and deep learning algorithms to comprehend and generate coherent text \cite{b1}. 
In the publishing industry, text review and editing is a critical process that involves the exploration and correction of manuscripts to enhance clarity, coherence, and grammatical accuracy \cite{b2}. 
The implementation of ChatGPT-4 in the editing of Spanish academic books could revolutionize the industry in several ways. 
Firstly, ChatGPT-4 can automate the correction of grammatical and spelling errors, increasing the efficiency of the editing process. 
Furthermore, its capability to analyze large volumes of text could facilitate the detection of inconsistencies and enhance coherence in lengthy texts \cite{b3}.

However, language models like ChatGPT-4 still face challenges in terms of contextual sensitivity and tone, necessitating a collaborative approach with human reviewers and editors to maintain quality \cite{b4}, \cite{b19}. 
When combined with human editors, ChatGPT-4 can enable a more agile and rigorous editing process, especially important in the academic context where accuracy and clarity are crucial. 
Additionally, ChatGPT-4 and similar models must be adequately trained and adapted to the particularities of the Spanish language, to ensure their effectiveness in editing texts in this language \cite{b5}.

This paper documents a research investigation that examined the performance of ChatGPT-4 in revising 100 academic books and compares the derived results with those produced by human reviewers (two for each book). The structure of this paper is organized as follows: Section \ref{RW} introduces studies similar to this research and provides a comparison. The applied methodology for this study is elaborated in Section \ref{M}. The results derived from this study are presented in Section \ref{R}, and a discussion regarding these results is conducted in Section \ref{D}. Lastly, chapter \ref{C} brings forth the conclusions drawn from this study and proposes potential future work in the same vein.

\section{Related work}\label{RW}

In the landscape of using AI for academic purposes, various studies have been conducted. One pertinent study in the realm of AI application in academic editing is the one conducted by Wang et al. \cite{b8}. They probed into the efficacy of ChatGPT for generating effective boolean queries in systematic literature reviews through multiple experiments. The results echoed the value of ChatGPT as a potent tool for conducting systematic reviews. However, they simultaneously highlighted challenges associated with striking a balance between automation and the critical analysis intrinsic to human-verified, documented work.

Checco et al., \cite{b9}, explore the possibility of employing artificial intelligence (AI) as a tool to aid or automate the process of peer-review. The authors construct a machine-learning mechanism, which is trained using 3300 conference papers, and the results exhibit that the system can aptly foresee the results of the peer-review procedure based on merely the manuscript's superficial traits. The study underscores the advantages of such AI applications, such as improved efficiency and enhanced understanding of the reviewing procedure. Nevertheless, it also emphasizes the requirement for addressing any prospective prejudices and ethical dilemmas related to these instruments.

In an examination of scientific manuscripts submitted to various artificial intelligence conferences, a comprehensive analysis was conducted. It comprised semantic, lexical, and psycholinguistic dissections of the manuscripts' full text, juxtaposed with the outcomes of the peer review process. The study determined that manuscripts that were accepted had lower readability scores and utilized more scientific and AI-related jargon than the rejected ones. Additionally, the accepted manuscripts employed less common, more abstract vocabulary, usually acquired at an older age. When analyzing the references within the manuscripts, it was observed that accepted submissions were more likely to cite identical publications—a fact reinforced by pairwise comparisons of manuscript word content, indicating greater semantic similarity among accepted works. Lastly, a prediction of the manuscripts' peer review outcomes was attempted based on their word content. The study concluded that usage of terms related to machine learning and neural networks was positively correlated with acceptance, whereas words tied to logic, symbolic processing, and knowledge-based systems were negatively correlated \cite{b10}.

Another recent work explored ChatGPT's potential role in peer review processes in academia, examining aspects such as the roles of reviewers and editors, review quality, reproducibility, and the social and epistemic implications of peer reviews. It was found that while LLMs like ChatGPT could assist in creating constructive reports and addressing review shortages, they also introduced potential biases, confidentiality issues, and concerns about reproducibility. Moreover, there are potential unforeseen social and epistemic consequences within academia due to partially delegating editorial work to LLMs. Despite the promise, the study calls for caution in using LLMs in academic contexts due to the possibility of amplifying existing biases and access inequalities \cite{b14}.

The study \cite{b15} investigated ChatGPT's competence in performing tasks related to bibliometric analysis, comparing the chatbot's outputs with those from a recent bibliometric study on a similar topic. Results demonstrated significant disparities and raised questions about the tool's reliability in this specific domain, thereby suggesting that researchers should proceed with caution when employing ChatGPT for bibliometric studies.

\section{Methodology}\label{M}

In contrast to related work, this study focuses on analyzing the similarities and differences between the content revisions of academic books conducted by ChatGPT-4 and human reviewers. This will help us determine the role this tool might play in the future of universities and literary publishing houses. To do this, we have selected a sample of 100 Spanish academic books that have been previously reviewed (some are already published) from January 2021 to June 2023 across 15 topics.

\subsection{Books selection}

Table \ref{books} displays a thematic classification of the books to be analyzed in this study. It is also noticeable that the demand for life science books surpasses that of other subjects. These areas of study are considered highly sensitive. Therefore, in these books, we have decided to place greater emphasis on the analysis of ChatGPT, but from a more ethical perspective.

\begin{table}
\center
\caption{Distribution of books analyzed by topic, average page count, and percentage of graphical content.}\label{books}
\begin{tabular}{|c|c|c|c|c|}
\hline
Topics & \begin{tabular}{@{}c@{}}Number \\of books\end{tabular} & \begin{tabular}{@{}c@{}}Average number \\ of pages \end{tabular} & \begin{tabular}{@{}c@{}}Percentage of \\ graphical content \end{tabular}\\
\hline
Advanced Mathematics & 5 & 150& 22 \% (mostly equations)\\
Architecture & 3 & 80&42 \%\\
Botany & 2 & 75&12 \%\\
Computer Sciences & 7 & 60&5 \% (code snipes)\\
Culture & 7 & 80&12 \% (mostly photographies)\\
Economics & 5 & 90& 6 \%\\
Education & 7 & 100& 5 \%\\
History & 4 & 150& 9 \%\\
Innovation & 3 & 40& 14 \%\\
Law and Legislation & 8 & 65& 4 \%\\
Medicine & 15  & 100& 13 \%\\
Odontology & 10 & 70& 42 \%\\
Pharmacology & 9 & 110& 37\% (mostly formulas)\\
Psychology & 10 & 130& 5 \%\\
Sociology & 5 & 120& 3 \%\\
\hline
\end{tabular}
\end{table}

\subsection{Preparation of the texts}

The text from these books will be prepared for processing. In certain instances, this involves digitizing the text if it is not already in a digital format, and ensuring that the text is free from any Optical Character Recognition (OCR) errors or any extraneous elements (such as watermarks) that might interfere with the text analysis \cite{b6}.

\subsection{Process of pair review and edition with reviewers and editors}

Each of these steps requires a careful, meticulous approach, and each plays a crucial role in ensuring that the final published book is accurate, engaging, and valuable to its intended audience. Note that depending on the size and structure of the publishing house, some or all of these tasks might be carried out by different individuals or even different departments.

Manuscript Assessment: In this first stage, the editor assesses the manuscript for its content, structure, and alignment with the intended audience. They provide an overall evaluation and make suggestions for improvement. This step often involves a lot of collaboration with the author.

Substantive or Developmental Editing: This is a deep, intensive editing phase focused on the content, structure, organization, and presentation of the manuscript. The editor may suggest reordering sections, rewriting passages, expanding or condensing content, and more. This stage requires a deep understanding of the subject matter.

Copy Editing: This is where the editor reviews the manuscript for grammar, punctuation, syntax, and consistency in style and voice. They will also check for consistency in facts and may cross-verify references and citations. The editor will ensure that the manuscript adheres to a specific style guide (like APA, Chicago, MLA, etc.).

Proofreading: This is the final stage of editing, which involves checking for typographical errors, spelling mistakes, and any inconsistencies missed during the copy-editing stage. The text is polished to be print-ready at this stage.

Indexing: While not editing per se, creating an index is an important part of preparing academic books. It involves listing the topics, names, and places mentioned in the book, along with page numbers, to help readers navigate the content.

Layout and Design: Although this is typically a designer's job, an editor might have some involvement in deciding the overall look and layout of the book, including the typography, cover design, and the arrangement of visual elements (like charts, graphs, or images).

Review: In academic publishing, the manuscript typically goes through a peer review process, where other experts in the field review the work for accuracy, relevance, and originality. The editor manages this process, communicates feedback to the authors, and oversees any revisions.

Approval and Publication: Once the manuscript has been thoroughly edited and reviewed, the editor gives their final approval for publication.

\subsection{Process of Review with ChatGPT-4}

\begin{figure}
\includegraphics[width=\textwidth]{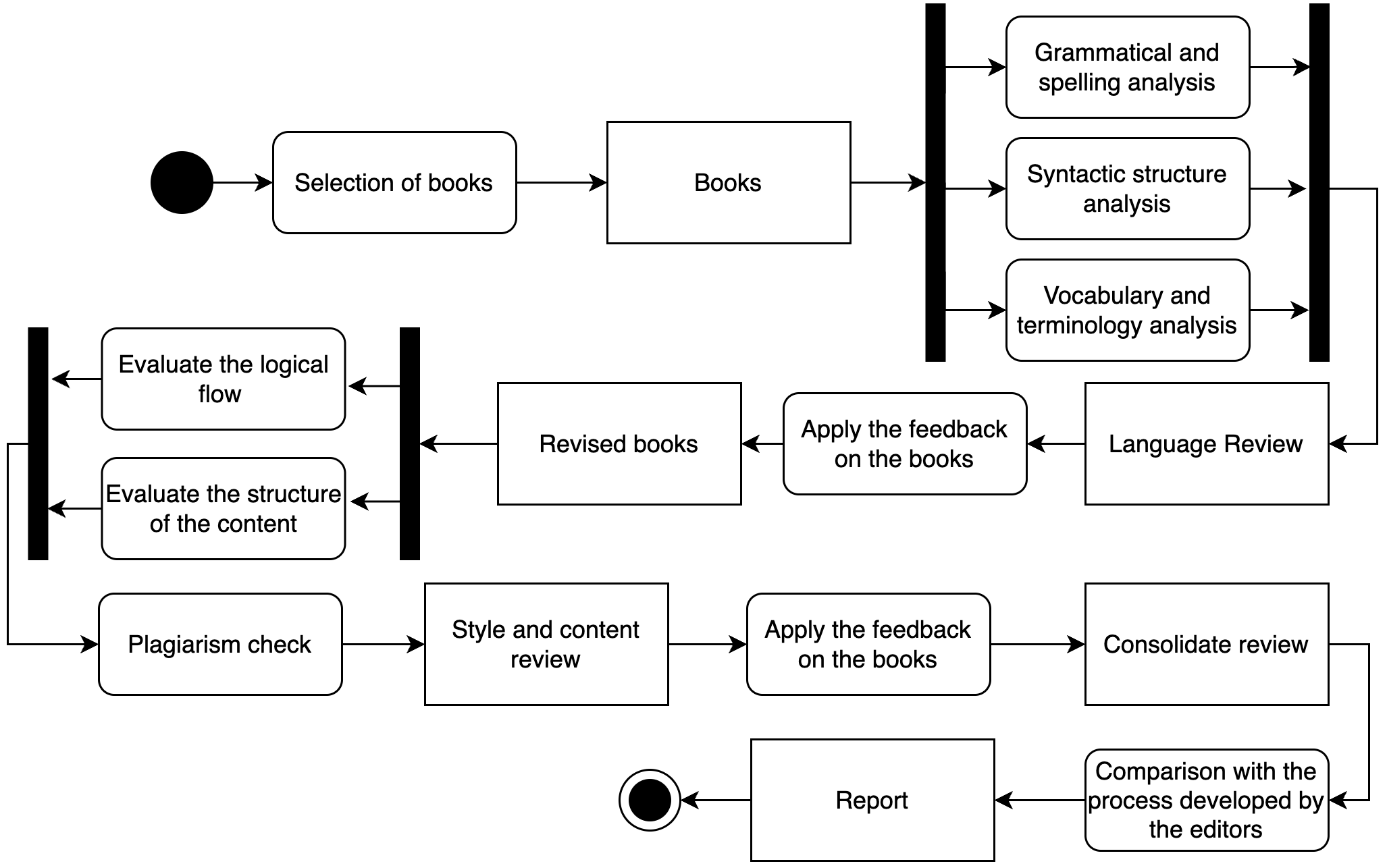}
\caption{Workflow to conduct the review of the selected books by ChatGPT before the comparison of results with the review made by the human reviewers and editors.} \label{flow}
\end{figure}

Each book was processed through ChatGPT-4 and the workflow is explained in Figure \ref{flow}. The AI conducted a comprehensive review and revision of the text, correcting any grammatical or spelling errors, and endeavored to improve the coherence and clarity of the text.
Furthermore, the AI also analyzed the syntactic structure of the sentences to ensure they conformed to grammatical rules. It assessed the use of tenses, punctuation, and other grammatical components for correctness and consistency throughout the text.

In addition to this, ChatGPT-4 analyzed the semantic aspects of the text. This included checking for the appropriate use of vocabulary and terminology, particularly important for subject-specific jargon in academic texts \cite{b7}.
The AI also aimed to evaluate the logical flow and structure of the content, checking for any inconsistencies or contradictions in the argument or presentation of the material.

Another critical aspect of this process was the evaluation of stylistic elements such as tone, voice, and overall readability of the text. While maintaining the original author's voice, the AI attempted to enhance the text's readability, ensuring that it was accessible and engaging for the intended audience.

In the next step, ChatGPT-4 performed an intertextual analysis, checking for potential plagiarism issues by comparing the text with a large corpus of academic literature. It also ensured correct citation and referencing as per the academic style guide relevant to each book.
Finally, a human editor will randomly select a book from each topic, analyze it with the assistance of ChatGPT, and then compare the time taken against another editor who will not use ChatGPT. The results are presented in Table 2.

\section{Results}\label{R}

In this study, we applied ChatGPT-4 to the task of editing 100 academic books in the Spanish language, covering a variety of academic fields. The results present a compelling case for the implementation of AI in the editorial process. Overall, ChatGPT-4 showed a high proficiency in identifying and correcting basic grammatical and spelling errors, demonstrating its value as a first-pass editing tool. 

Table \ref{resultsTable} displays the average time that ChatGPT required to analyze the text of the selected book collection. It's evident that topics primarily composed of text require less time for ChatGPT to interpret, and the total number of errors is lower compared to books with more graphical content like tables, figures, formulas, and equations.

\begin{table}
\center
\caption{Average time spent by ChatGPT to analyze the book collection, total errors recorded in the books by topic, and the time saved for a human editor when he collaborated with ChatGPT to analyze one random book of each topic.}\label{resultsTable}
\begin{tabular}{|c|c|c|c|c|cl}
\hline
Book Topics & \begin{tabular}{@{}c@{}}ChatGPT \\invested time\end{tabular} & \begin{tabular}{@{}c@{}}Misunderstandings and \\ errors by ChatGPT\end{tabular}  & Saved time\\
\hline
Advanced Mathematics & 88 min & 142 (5 books) & 9 \%\\
Architecture & 56 min & 35 (3 books)& 8 \%\\
Botany & 49 min & 12 (2 books)& 32 \%\\
Computer Sciences & 328 min & 28 (7 books)&14 \%\\
Culture & 252 min & 27 (7 books)& 25 \%\\
Economics & 194 min & 28 (5 books)& 38 \%\\
Education & 132 min & 29 (7 books)& 45 \%\\
History & 115 min & 31 (4 books)& 51 \%\\
Innovation & 95 min & 54 (3 books)& 26 \%\\
Law and Legislation & 74 min & 65 (8 books)& 47 \%\\
Medicine & 977 min  & 58 (15 books)& 37 \%\\
Odontology & 630 min & 43 (10 books)& 12 \%\\
Pharmacology & 742 min & 32 (9 books)& 29 \%\\
Psychology & 257 min & 23 (10 books)& 47 \%\\
Sociology & 139 min & 21 (5 books)& 55 \%\\
\hline
\end{tabular}
\end{table}

We split the results by analyzing limitations and the qualitative comparison with a human editor explaining advantages and disadvantages and finally, we pointed out in which sense ChatGPT and a human editor are similar.

\subsection{Limitations}

Although ChatGPT is a powerful tool, there are still several limitations to be taken into account when using it for book reviewing \cite{b12}, \cite{b13}, \cite{b17}. These limitations have been considered in some approaches to avoid academic misconduct \cite{b18}. 

\begin{enumerate}
\setlength{\itemsep}{0cm}
\setlength{\parskip}{0cm}
\item \textbf{Token Limitation:} ChatGPT has a limit to the number of tokens (units of text) it can process at one time. For GPT-4, this limit is 2048 tokens. Therefore, it may not be able to analyze very long texts in their entirety in a single pass. This was time-consuming afterward because we had to extract only the text from all the collections of books. 
\item \textbf{Inability to Interpret Visual Elements:} ChatGPT is a natural language processing model and cannot analyze or interpret visual elements such as graphs, tables, images, or diagrams. This means it cannot verify the accuracy or relevance of such elements within the text, which results in several misunderstandings in books that had a considerable amount of this content (e.g. Advanced Mathematics, Architecture, Odontology).
\item \textbf{Inability to Access External Databases in Real-Time:} ChatGPT cannot access external databases or browse the internet in real-time to verify facts or references. Its knowledge is based solely on the data it was trained with and does not update with new or real-time information.
\item \textbf{No File System Interactivity:} ChatGPT cannot interact with a computer or network's file system, meaning it cannot open, close, read, or write files directly. This must be done by a human to correct the text that was reviewed by ChatGPT. 
\item \textbf{Inability to Interpret Markup Languages:} ChatGPT may struggle to interpret and work with markup languages such as HTML, LaTeX, or XML which are often used in the production of academic documents and books. This must be done by a human and correct the text that was reviewed by ChatGPT on the source files. 
\end{enumerate}

\subsection{Qualitative comparison of the results of the review with ChatGPT-4 with the review by human reviewers}

ChatGPT, as an artificial intelligence language model, can offer certain advantages compared to human reviewers of academic books. However, it's important to note that it does not entirely replace human editing and reviewing but can complement it. The key advantages of ChatGPT compared to a human editor include: 

\begin{enumerate}
\setlength{\itemsep}{0cm}
\setlength{\parskip}{0cm}
  \item \textbf{Efficiency:} ChatGPT can process and analyze large volumes of text in a short period, which can enhance efficiency in reviewing lengthy documents. 
  Even though books with many graphical elements combined with text took longer for ChatGPT to analyze and subsequently required correction by the human editor, there is still a noticeable time-saving benefit (e.g. Advanced Mathematics - 9\% time saved). 

  \item \textbf{Availability:} As an artificial intelligence program, ChatGPT is available 24/7, does not need breaks or time off, and can work at the same constant speed for extended periods.
  \item \textbf{Basic error correction:} ChatGPT can swiftly detect and correct grammatical, orthographic, and punctuation errors. However, its effectiveness depends on the language and the specific characteristics of the text.
  \item \textbf{Consistency:} ChatGPT can maintain consistent coherence in the application of grammatical and stylistic rules, which is particularly useful for lengthy documents where maintaining consistency can be a challenge for human reviewers.  
\end{enumerate}

While artificial intelligence models like ChatGPT have made remarkable strides in language processing tasks \cite{b11}, there are still several areas where human reviewers and editors outperform AI when it comes to reviewing academic books:

\begin{enumerate}
\setlength{\itemsep}{0cm}
\setlength{\parskip}{0cm}
\item \textbf{Contextual Understanding: } Human reviewers can better understand complex contextual information that AI might miss. They can interpret nuances, subtleties, and complexities within the text that an AI might not fully comprehend.

\item \textbf{Style and Tone:} Humans have a deeper understanding of the author's style and tone, which is critical for preserving the author's unique voice in academic writing. Humans can also detect and correct inappropriate shifts in tone more effectively than AI.

\item \textbf{Logic and Argumentation:} Human reviewers can evaluate the logical flow of arguments and the presentation of ideas, which is vital in academic books. They can recognize whether the author's argumentation is solid, consistent, and properly substantiated.

  \item \textbf{Subjective Judgement:} Humans can make subjective judgments that AI can't. This includes evaluating whether the text is engaging, interesting, or appropriate for the intended audience.

\item \textbf{Ethical Considerations:} Human reviewers are better at detecting ethical issues in the text, such as cultural insensitivity, bias, or potential plagiarism, that may not be readily apparent to AI. 

\item \textbf{Adherence to Guidelines:} Human reviewers can ensure adherence to specific academic or publishing guidelines, which may involve complex or subjective elements that AI could struggle to interpret. 
\end{enumerate}

\subsection{Quantitative comparison of the results of the review with ChatGPT-4 with the review by human reviewers}

In future versions, two areas where ChatGPT should excel would be:
\begin{enumerate}

\item \textbf{Processing of multiple documents:} If ChatGPT could handle multiple documents simultaneously, it would have a significant advantage over a human editor. At the moment both have this limitation.

\item \textbf{Style and voice:} The human editor acknowledges that identifying and preserving an author's unique style and voice can be challenging. ChatGPT also faces this challenge in certain cases.
  
\end{enumerate}

\section{Discussion}\label{D}
 
Table \ref{resultsTable} illustrates that based on the content percentage (text, figures, tables, equations, and formulas) and the length of the book, ChatGPT can be immensely valuable for various stages of a traditional editorial process.

A book containing a mix of text with images, equations, tables, or formulas can be confusing for ChatGPT, especially when the text references content that ChatGPT cannot read. Therefore, for primarily text-based books, ChatGPT is faster and makes fewer errors (e.g. Law and Legislation, Psychology and Sociology). This means a human editor would spend less time addressing these misunderstandings that ChatGPT encounters.

\begin{figure}
\includegraphics[width=\textwidth]{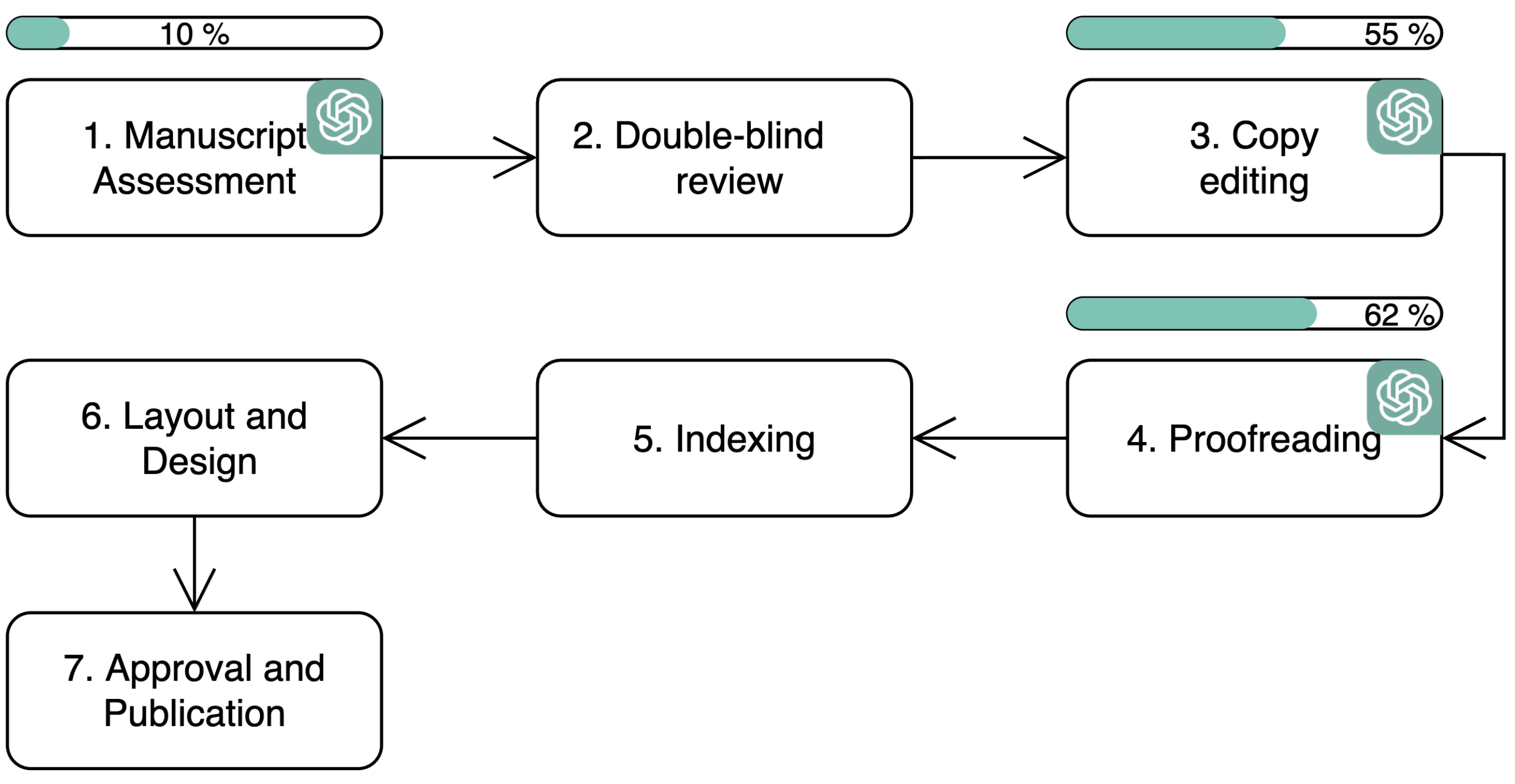}
\caption{Summary of the editorial process stages where ChatGPT can assist human editors and the degree of its potential contribution.} \label{fig1}
\end{figure}

Based on the results Table \ref{resultsTable}, Figure \ref{fig1} provides a summarization of the specific stages in the editorial process where ChatGPT can lend support to human editors, along with the extent of its potential contribution. 

The calculation of how beneficial ChatGPT can be in subsequent stages is based on the average time it took a human editor to review a randomly selected book from each topic using ChatGPT, compared to the time it took another human editor to perform the same task without ChatGPT.

\begin{itemize}
\setlength{\itemsep}{0cm}
\setlength{\parskip}{0cm}
\item \textbf{1. Manuscript Assessment: } ChatGPT could potentially assist in conducting an initial assessment of a manuscript by providing a general analysis of the text, including measures of readability, stylistic metrics, and possible inconsistencies in tense, style writing, or voice (in the best case). It could also help in identifying overtly complex sections that may need to be rewritten for clarity, which could give editors a high-level view of the manuscript's quality and make a decision about its possible publication. However, ChatGPT cannot replace an editorial committee. Depending on their policies, collections, and audience, the committee will decide on the relevance and potential for publication.

\item \textbf{3. Copy Editing: } This is perhaps the area where ChatGPT could offer the most assistance to editors. The AI could be used to spot and correct issues related to grammar, punctuation, and syntax. It could also help ensure consistency in terms of the use of terminologies across the manuscript. However, this varies depending on the book. In multi-author books, it's challenging for both ChatGPT and a human editor to adjust the style and voice to a single standard.

\item \textbf{4. Proofreading:} ChatGPT could serve as an effective tool for spotting typographical errors and minor mistakes that may have been overlooked during the previous editing stages. Given its ability to process large amounts of text quickly, it could be especially useful for the final proofreading of lengthy academic texts. However, at this stage, it's crucial to provide specific instructions to only analyze these types of errors. Otherwise, ChatGPT begins to overemphasize its observations regarding style and voice, which can lead to a cycle of corrections upon corrections.

\end{itemize}

While ChatGPT can significantly aid some parts of the editorial process (Figure \ref{fig1}), it doesn't replace the need for a human editor or reviewer. The AI tool lacks a deep understanding of context, cannot appreciate the nuances of language in the same way a human can, and does not possess the ability to fact-check content or verify the validity of the arguments presented with the references.

\begin{itemize}
\setlength{\itemsep}{0cm}
\setlength{\parskip}{0cm}

\item \textbf{2. Double-blind review: }The peer review process involves managing the exchange of complex, specialized feedback between experts in a specific field. While ChatGPT can generate text and summaries, it lacks the deep, field-specific knowledge and human judgment required to understand and manage this feedback. It also can't ensure that the feedback has been adequately addressed in revisions.

\item \textbf{5. Indexing: }Creating an index for a book involves understanding the core themes, concepts, and key figures within a text and associating them with specific page numbers. While ChatGPT can identify keywords and themes, it lacks an understanding of the relative importance of these elements within the broader context of the book. Moreover, it is currently incapable of interacting with pagination information, which is crucial for this task.

\item \textbf{6. Layout and Design: }This stage involves aesthetic judgments and an understanding of visual design principles, which are outside the capabilities of a text-based AI model like ChatGPT. It can't process or make decisions about visual elements such as typography, cover design, or the arrangement of charts, graphs, or images. And even this is a problem when the text refers to a figure or table. ChatGPT just makes some assumptions and this leads to bad analysis. 

\item \textbf{7. Approval and Publication: }The final approval for publication involves making judgments on the manuscript as a whole. This requires a holistic understanding of the content, context, and relevance of the work, which is beyond the capabilities of current AI models. This stage may also involve tasks such as communicating with publishing services and making decisions about publication timelines, tasks that AI models like ChatGPT are not currently equipped to handle.
\end{itemize}

Based on the evidence gathered throughout this study, coupled with insights derived from an extensive examination of relevant literature, it is evident that ChatGPT-4, albeit a highly advanced AI language model, should be optimally deployed as an augmentative instrument to enhance the capabilities of human editors, rather than serve as a complete replacement. Its efficacy primarily lies in its ability to support and streamline certain aspects of the editing process, while crucial tasks that require nuanced understanding, holistic judgment, and sophisticated design aesthetics continue to necessitate the unique expertise of human professionals.

One of the editors involved in this study highlighted a prevalent issue where ChatGPT may not offer much assistance. "Spanish, like many languages in Latin America, is replete with regionalisms. Often, it is necessary to engage with the author directly to clarify their intended meaning without altering the essence of their idea, thereby preserving the context of their research".

"At times, it's more advantageous to first analyze texts from the perspective of their region of origin before neutralizing them and enhancing clarity. This scenario occurs especially in social sciences or art books, where authors take more liberty in describing situations, artworks, or investigations".

"Another aspect that often slows down the editing process is the presence of formulas and symbols, common in mathematical, physical, or chemical texts. Since an editor may not specialize in these precise scientific branches, they often need to return this portion of the text to the author for verification of data before finalizing the editing process". 

This complex interplay between contextual understanding, content authenticity, and technical validation, further underscores the limitations of AI tools like ChatGPT and the indispensable role of human reviewers and editors. 

\section{Conclusions}\label{C}

ChatGPT undoubtedly offers the potential for automating aspects of the academic book-reviewing process, but its technical limitations underscore the continuing necessity for human involvement. The model excels at preliminary editing and proofreading tasks, such as correcting grammar and spelling mistakes, and making basic stylistic improvements. However, the indispensable role of human oversight in the process can't be overlooked.

Despite ChatGPT's powerful capabilities, it faces substantial challenges such as limited deep contextual understanding, interpretation of subtle nuances, and the absence of human-like editorial judgment. Consequently, an AI-edited text might not match the standards of one edited by a human. Moreover, the model's inability to interact with visual content like graphs and tables, process real-time data, or understand markup languages further cements the need for human editors.

Another area of concern is the model's ability to handle lengthy texts consistently, treat sensitive content appropriately, and prevent the propagation of errors. These issues are difficult for AI to manage and they underline the requirement for a human touch in the editing process.

AI tools such as ChatGPT have further limitations when used in bibliometric analysis, as evidenced by comparisons with traditional bibliometric studies. Discrepancies between results indicate that trustworthiness and reliability in such a context are still not at the desired levels.

These limitations are also shared by other Large Language Models (LLMs) such as Bard. However, Bard could potentially assist in verifying the truthfulness of a book's content \cite{b16}. This is a promising area of investigation that we plan to explore in future research. A possible approach would be to compare how 'Bard' or 'Llama 2' respond to this same experiment and highlight the comparison of the results between these LLMs.

In conclusion, while AI exhibits substantial potential in aiding the editorial process, it is not yet poised to fully replace human editors. The optimal approach may involve a synergistic relationship between AI and human reviewers, integrating the strengths of both to ensure high-quality, contextually nuanced, and ethically sound editing in academic books. As we move forward, the evolving landscape of AI in academia will necessitate careful consideration of how to effectively balance human and AI contributions to ensure the highest standards of academic integrity.

\subsubsection{Acknowledgements} The authors of this work would like to thank the CEDIA editorial team for allowing them to conduct this research and for welcoming the analysis of this Artificial Intelligence tool with the works that have been reviewed since the inception of the editorial.


\begin{thebibliography}{8}
\bibitem{b1}
Brown, T., Mann, B., Ryder, N., Subbiah, M., Kaplan, J. D., Dhariwal, P., ..., \& Amodei, D.: Language models are few-shot learners. Advances in Neural Information Processing Systems \textbf{33}, 1877--1901 (2020).

\bibitem{b2}
Einsohn, A., \& Schwartz, M.: The copyeditor's handbook: A guide for book publishing and corporate communications. 4th edn. University of California Press, USA (2019).

\bibitem{b3}
Indurkhya, N., \& Damerau, F. J. (Eds.): Handbook of natural language processing. Vol. 2. CRC Press, (2010).

\bibitem{b4}
Bender, E. M., \& Koller, A.: Climbing towards NLU: On meaning, form, and understanding in the age of data. In: Proceedings of the 58th Annual Meeting of the Association for Computational Linguistics, pp. 5185--5198. (2020).

\bibitem{b19}Kasneci, E., Seßler, K., Küchemann, S., Bannert, M., Dementieva, D., Fischer, F., ... \& Kasneci, G. (2023). "ChatGPT for good? On opportunities and challenges of large language models for education." Learning and Individual Differences, 103, 102274. 

\bibitem{b5}
Tiedemann, J., \& Agić, Z.: Synthetic treebanking for cross-lingual dependency parsing. Journal of Artificial Intelligence Research \textbf{55}, 209.

\bibitem{b6}
Wu, J., Ouyang, L., Ziegler, D. M., Stiennon, N., Lowe, R., Leike, J., \& Christiano, P.: Recursively summarizing books with human feedback. arXiv preprint \textbf{2109.10862} (2021).

\bibitem{b7}
Fang, T., Yang, S., Lan, K., Wong, D. F., Hu, J., Chao, L. S., \& Zhang, Y.: Is chatgpt a highly fluent grammatical error correction system? A comprehensive evaluation. arXiv preprint \textbf{2304.01746} (2023).

\bibitem{b8}
Wang, S., Scells, H., Koopman, B., \& Zuccon, G.: Can ChatGPT write a good boolean query for systematic review literature search?. arXiv preprint \textbf{2302.03495} (2023).

\bibitem{b9}
Hyland-Wood, B., Gardner, J., Leask, J., \& Ecker, U. K.: Toward effective government communication strategies in the era of COVID-19. Humanities and Social Sciences Communications \textbf{8}(1) (2021).

\bibitem{b10}
Vincent-Lamarre, P., \& Larivière, V.: Textual analysis of artificial intelligence manuscripts reveals features associated with peer review outcome. Quantitative Science Studies \textbf{2}(2), 662--677 (2021)

\bibitem{b11}
Bukar, U., Sayeed, M. S., Razak, S. F. A., Yogarayan, S., \& Amodu, O. A.: Text analysis of chatgpt as a tool for academic progress or exploitation. \url{http://ssrn.com/abstract=4381394}. Last accessed (July, 2023)

\bibitem{b12}
OpenAI: GPT-4 Technical Report. arXiv preprint \textbf{2303.08774} (2023)

\bibitem{b13}
Tamkin, A., Brundage, M., Clark, J., \& Ganguli, D.: Understanding the capabilities, limitations, and societal impact of large language models. arXiv preprint \textbf{2102.02503} (2021)

\bibitem{b14}
Hosseini, M., \& Horbach, S. P.: Fighting reviewer fatigue or amplifying bias? Considerations and recommendations for use of ChatGPT and other Large Language Models in scholarly peer review. Research Integrity and Peer Review \textbf{8}(1), 4 (2023).

\bibitem{b15}
Farhat, F., Sohail, S. S., \& Madsen, D. Ø.: How trustworthy is ChatGPT? The case of bibliometric analyses. Cogent Engineering \textbf{10}(1), 2222988 (2023).

\bibitem{b16}
Siad, S. M.: The Promise and Perils of Google's Bard for Scientific Research. (2023).

\bibitem{b17}
Hill-Yardin, E. L., Hutchinson, M. R., Laycock, R., \& Spencer, S. J.: A Chat (GPT) about the future of scientific publishing. Brain Behav Immun \textbf{110}, 152--154 (2023).

\bibitem{b18}
Berrezueta-Guzman, J., Krusche, S.: Recommendations to Create Programming Exercises to Overcome ChatGPT, edarxiv.org/expuq.



\end{thebibliography}
\end{document}